\documentclass[conference]{IEEEtran}
\IEEEoverridecommandlockouts

\usepackage{cite}
\usepackage{amsmath,amssymb,amsfonts}
\usepackage{algorithmic}
\usepackage{graphicx}
\usepackage{textcomp}
\usepackage{xcolor}
\usepackage{url}
\usepackage{hyperref}

\usepackage{graphicx}
\usepackage{newtxmath}
\usepackage{multirow}
\usepackage{booktabs}
\usepackage{CJK}
\usepackage{xpinyin}
\usepackage{color}

\newcommand{\methodName}{CL}

\def\BibTeX{{\rm B\kern-.05em{\sc i\kern-.025em b}\kern-.08em
    T\kern-.1667em\lower.7ex\hbox{E}\kern-.125emX}}

\begin{document}
\title{Loss-Aware Curriculum Learning for Chinese Grammatical Error Correction}

\author{\IEEEauthorblockN{1\textsuperscript{st} Ding Zhang$^{\dagger}$}
\IEEEauthorblockA{\textit{Shenzhen International Graduate School} \\
\textit{Tsinghua University}\\
Shenzhen, China \\
zhangd22@mails.tsinghua.edu.cn}
\and
\IEEEauthorblockN{2\textsuperscript{nd} Yangning Li$^{\dagger}$}
\IEEEauthorblockA{\textit{Shenzhen International Graduate School} \\
\textit{Tsinghua University}\\
Shenzhen, China \\
yn-li23@mails.tsinghua.edu.cn}
\and
\IEEEauthorblockN{3\textsuperscript{rd} Lichen Bai}
\IEEEauthorblockA{\textit{Shenzhen International Graduate School} \\
\textit{Tsinghua University}\\
Shenzhen, China \\
blc22@mails.tsinghua.edu.cn}
\and
\IEEEauthorblockN{4\textsuperscript{th} Hao Zhang}
\IEEEauthorblockA{\textit{Shenzhen International Graduate School} \\
\textit{Tsinghua University}\\
Shenzhen, China \\
zhang-h22@mails.tsinghua.edu.cn}
\and
\IEEEauthorblockN{5\textsuperscript{th} Yinghui Li}
\IEEEauthorblockA{\textit{Shenzhen International Graduate School} \\
\textit{Tsinghua University}\\
Shenzhen, China \\
liyinghu20@mails.tsinghua.edu.cn}
\and
\IEEEauthorblockN{6\textsuperscript{th} Haiye Lin}
\IEEEauthorblockA{\textit{Shenzhen International Graduate School} \\
\textit{Tsinghua University}\\
Shenzhen, China \\
lin-hy22@mails.tsinghua.edu.cn}
\and
\IEEEauthorblockN{7\textsuperscript{th} Hai-Tao Zheng$^{*}$}
\IEEEauthorblockA{\textit{Shenzhen International Graduate School} \\
\textit{Tsinghua University, Peng Cheng Laboratory}\\
Shenzhen, China \\
zheng.haitao@sz.tsinghua.edu.cn}
\and
\IEEEauthorblockN{8\textsuperscript{th} Xin Su}
\IEEEauthorblockA{\textit{WeChat} \\
\textit{Tencent}\\
Shenzhen, China \\
levisu@tencent.com}
\and
\IEEEauthorblockN{9\textsuperscript{th} Zifei shan}
\IEEEauthorblockA{\textit{WeChat} \\
\textit{Tencent}\\
Shenzhen, China \\
zifeishan@tencent.com}
\thanks{$^{\dagger}$ Equally contribution
}
\thanks{* Corresponding author. (E-mail: zheng.haitao@sz.tsinghua.edu.cn) This research is supported by National Natural Science Foundation of China(Grant No.62276154), Research Center for ComputerNetwork (Shenzhen) Ministry of Education, the Natural Science Foundation of Guangdong Province(Grant No.2023A1515012914 and 440300241033100801770), Basic Research Fund of Shenzhen City (Grant No.JCYJ20210324120012033, JCYJ20240813112009013 and GJHZ20240218113603006), the Major Key Projectof PCL for Experiments and Applications (PCL2023A09).
}
}




\maketitle

\begin{abstract}
Chinese grammatical error correction (CGEC) aims to detect and correct errors in the input Chinese sentences. Recently, Pre-trained Language Models (PLMS) have been employed to improve the performance. However, current approaches ignore that correction difficulty varies across different instances and treat these samples equally, enhancing the challenge of model learning. To address this problem, we propose a multi-granularity Curriculum Learning (CL) framework. Specifically, we first calculate the correction difficulty of these samples and feed them into the model from easy to hard batch by batch. Then Instance-Level CL is employed to help the model optimize in the appropriate direction automatically by regulating the loss function. Extensive experimental results and comprehensive analyses of various datasets prove the effectiveness of our method.
\end{abstract}

\begin{IEEEkeywords}
Pre-trained Language Models, Chinese Grammatical Error Correction, Curriculum Learning\end{IEEEkeywords}

\section{Introduction}
The task of Chinese Grammatical Error Correction (CGEC) involves automatically identifying and correcting grammatical mistakes in the input Chinese sentences~\cite{li2024correct}. CGEC has attracted increasing attention from NLP researchers because it benefits many applications, such as Writing Assistant
~\cite{li2023towards}, Automatic Speech Recognition~\cite{banno2023towards}, and Search Engine~\cite{duan2011online}.

Recently, methods based on Pre-trained Language Models have become the mainstream of the CGEC task~\cite{li2023effectiveness,li2022past}. Existing methods based on PLMs for CGEC can be categorized into three categories: Sequence-to-Sequence method(Seq2seq)~\cite{ye2023cleme}, Sequence-to-Edit method(Seq2Edit)~\cite{ma2022linguistic}, and Ensemble models~\cite{li2022learning}.
The Seq2Edit method treats CGEC as a sequence labeling task and adopts editing operations between the source text and the target text as training objectives. The Seq2Seq method treats the CGEC problem as a monolingual translation problem, where the erroneous text is regarded as the source text and the correct sentence serves as the target text, respectively. Ensemble models combine the two methods described above and can easily fix more error types.

However, the current approach faces a severe problem in that it ignores the diversity of CGEC training data, where correction difficulty varies across different instances~\cite{li-etal-2023-grammatical,lichtarge-etal-2020-data}. Equally treating these training data would have a negative influence on the performance of CGEC models.





The core idea of Curriculum Learning (CL) is to facilitate the model's training process by learning from simple samples to complex ones, which has demonstrated
its efficacy in improving the performance of natural machine translation systems~\cite{bengio2009curriculum,lu-zhang-2021-exploiting-curriculum,li2024llms,li2023active,li2024ecomgpt}. Therefore, we introduce a multi-granularity Curriculum Learning framework to address the above issue in this paper. Intuitively, hard samples should be difficult for CGEC models to correct. So we first use the training loss to measure the difficulty of each pair of erroneous sentences and correct sentences. Then samples will be fed into CGEC models batch by batch in ascending order based on their difficulty scores. 

However, the Batch-Level CL method is rough, and the difficulty scores of training data at the Sentence-Level or Token-Level within a specific batch can vary, an issue that also needs to be addressed. To achieve such fine-grained learning, we determine to assign different weights to simple and complex samples by regulating the loss function.  Specifically, we first employ the model to evaluate the quality scores of instances in the training set. Then the scores serve as learning factors to dynamically adjust the contributions of different parts when computing the training loss, encouraging the CGEC model to focus on the portions that are difficult to correct.

To verify the efficacy of our framework, we conduct extensive
experiments using PLMs such as mT5 and BART. The experimental results demonstrate that our proposed Curriculum Learning method outperforms the baseline scores on NLPCC~\cite{zhao2018overview} and MuCGEC~\cite{zhang2022mucgec} datasets, proving that our framework can boost current CGEC models.
The contributions of this paper can be summarized as follows:
\begin{itemize}
\item We develop a multi-granularity Curriculum Learning method to tackle the CGEC task.
\item By utilizing the difficulty of training data from multi-granularity, our method enables CGEC to focus on the more challenging aspects of the data and optimize in the appropriate direction.
\item Extensive experiments verify that our proposed CL method is able to improve the performance of the CGEC models based on PLMs.
\end{itemize}

\section{Proposed approach}
In this section, we will first review the fundamental definition of the Chinese grammatical error correction task. Then we will introduce our proposed CL framework for CGEC, which comprises two sub-modules operating at distinct levels.

\subsection{Problem Definition}
\label{sec:Definition}
Let $x^{i} = \left[x^{i}_{1},x^{i}_{2},...,x^{i}_{n} \right]$ denotes the i-th source sentence and $y^{i} = \left[y^{i}_{1},y^{i}_{2},...,y^{i}_{m} \right]$ is its corresponding corrected sentence. CGEC employs a pre-trained language model to build a correction model, aiming to maximize the conditional distribution of error-corrected sentence pairs in a parallel corpus. More specifically, the learning objective is to minimize negative log-likelihood loss:
\begin{equation}
   L_{(x^{i},y^{i})} = -\frac{1}{m}\sum^{m}_{t=1}logP(y^{i}_{t} |y^{i}_{<t}, x^{i};\theta)  \label{eq:cross-loss}
\end{equation}
where $\theta$ denotes the trainable parameters of our CGEC model.

\begin{figure}[]
\centering
\small
\includegraphics[width=0.45\textwidth]{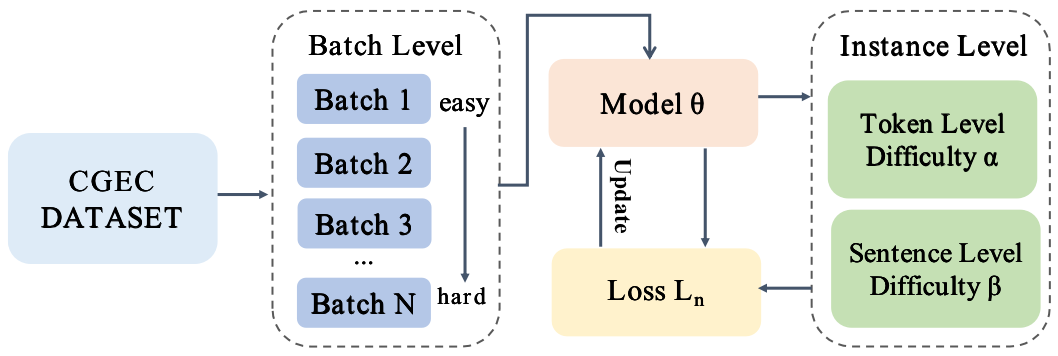}
\caption{An Overview of Loss-Aware Curriculum Learning (CL) framework for the CGEC task.}
\label{Method_Figure}
\end{figure}

As shown in Fig~\ref{Method_Figure}, the essence of our method lies in how to enforce CGEC models to pay more attention to complex samples.

\subsection{Batch-Level Curriculum Learning}
\label{sec:batch_cl}
\subsubsection{Difficulty Criteria}
We first use the cross-entropy loss function to measure the difficulty of each data sample. Then we will control the dataloader to load samples of the training set from simple to complex at the Batch-Level. 

As mentioned above, evaluating data difficulty is essential for Curriculum Learning~\cite{wang-etal-2019-dynamically,liu-etal-2020-norm}. Traditional difficulty criteria, such as word frequency or sentence length, are not able to reflect the intrinsic difficulty of correcting error text pairs.
Based on detailed observation, we find that the CGEC model can easily correct the error of data with a lower loss but is not good at correcting the error of data with a higher loss. Therefore, we first propose to determine the data difficulty according to the cross-entropy loss function in the CGEC task.

Given a trained CGEC model and a dataset with N sentence pairs $\{(x^1,y^1),(x^2,y^2),...,(x^N,y^N)\}$, we can acquire the difficulty of each instance by calculating the cross-entropy loss function. Then Cumulative Density Function (CDF) is employed to transfer the distribution of instances difficulty into
(0, 1]:
\begin{equation}
   \small
   \widetilde{d}\left<(x^i,y^i)\right> \in (0,1] = CDF(\{d\left<(x^i,y^i)\right>\}_{i=1}^{N})^i \label{eq:diff-cdf}
\end{equation}
Specifically, the score of the difficult instance tends to be 1, while that of the easy instance tends to be 0.
\subsubsection{Curriculum Arrangement}
The second question in Curriculum Learning is how to organize all the
training samples into a sequenced curriculum based on their difficulty
scores, which determines the complexity of samples the CGEC model can acquire at a specific batch. 

Following ~\cite{platanios-etal-2019-competence}, we use the notion of model competence, which is a function that takes the step $t$ of a training batch as input and outputs a value from 0 to 1 to control the loading schedule of training data in Batch-Level.

\begin{equation}
    C(t)=min\left(1,\sqrt[k]{{t\cdot\frac{1-c_{0}^k}{ T}}+c_{0}^k}\right)  \label{eq:root-q}
\end{equation}

where $c_0$ is the initial competence at the beginning of training,
k is the coefficient to control the growth rate of model competence, 
T is a hyper-parameter to determine the length of the Batch-Level curriculum.

At step $t$, all the available samples whose loss difficulty scores are less than $c_t$ are available to the CGEC model. Through the learning strategy, the CGEC model progressively acquires the training samples from simple to complex batch by batch.

\subsection{Instance-Level Curriculum Learning}
\label{sec:instance_cl}

While Batch-Level CL is able to enhance the CGEC model by controlling the dataloader to load training data from easy samples to hard ones sequentially, it is important to note that the qualities and levels of difficulty among sentences and tokens in a given batch are also different. Therefore, we propose to determine the learning emphasis in a particular batch according to the instance difficulty. 

Intuitively, common tokens tend to be easier for the model to generate, while rare tokens tend to be difficult during the generation.
As the length of the sentences and the quantity of grammatical mistakes both increase, the complexity of fixing the erroneous text also escalates. Therefore, we use Monte Carlo dropout sampling to calculate the conditional probabilities so as to modify the loss weights of the simple instances and the complex instances dynamically.

Given the current CGEC model parameterized by $\theta$ and a mini-batch consisting of M sentence pairs 
$\{(x^1, y^1), (x^2, y^2)\\, ..., (x^M, y^M)\}$, the number of Q Monte Carlo dropout sampling is employed on the CGEC model.
Thus, each instance is able to yield Q conditional probabilities. The variance of the probabilities can reflect the confidence that the model has with respect to each instance. As a result, we utilize the probabilities to assess the complexity of each individual sentence and token. 

\section{Experiments}
\subsection{Datasets}
\label{ssec:dataset}

Following previous works, we evaluate the model performance on two recent CGEC datasets, NLPCC and MuCGEC.
Lang8~\cite{zhao2018overview} and HSK~\cite{zhang2009features} are used for model training. Lang8 is collected from a language learning website and HSK is collected from official Chinese proficiency tests.  The details of the datasets are provided in Table~\ref{Data_Statistics}.

\begin{table}[htbp]
\caption{Statistics of datasets, including the number of paired samples(Num), the average length of per source sentence(Avg. Length), the number of erroneous samples(Err. Sample)}
\small
\centering
\begin{tabular}{lrrr}
\hline Training Data & Num & Avg. Length & Err. Sample \\
\hline Lang8 & 1,220,906 & $18.9$ & 1,096,387 \\
HSK & 156,870 & $27.3$ &  96,049 \\
\hline Total & 1,377,776 & $19.8$ & 1,192,436 \\
\hline 
\\
\hline Test Data & Num & Avg. Length & Err. Sample \\
\hline NLPCC-test & 2,000 & $29.7$ & 1,224 \\
MuCGEC-dev & 2,467 & $44.0$ &  2,412 \\
\hline
\end{tabular}

\label{Data_Statistics}
\end{table}

\subsection{Evaluation Metrics}
\label{ssec:Metrics}
For the NLPCC dataset, we present the results measured
by the $M^{2}$ scorer for evaluation. For the MuCGEC set, ChERRANT\footnote{\url{https://github.com/HillZhang1999/MuCGEC/tree/main/scorers/ChERRANT}} scores are employed to evaluate the performance. 
All our results are an average of five distinct experiments using different random seeds.


\begin{CJK*}{UTF8}{gbsn}

\begin{table}[ht]
\small
\centering
\caption{Zero-shot settings for GPT-3.5-Turbo and GPT-4.}

\resizebox{0.46\textwidth}{!}{
\begin{tabular}{p{1.9cm} | p{6cm}}
\toprule
 \multicolumn{1}{c|}{\textbf{Method}} &  \multicolumn{1}{c}{\textbf{CGEC Prompt}} \\
\cmidrule(lr){1-2}
\multirow{8}{*}{\textbf{Zero-shot}} &\texttt{"role": "system", "content": "您是一个优秀的中文语言学家，您的职责是识别并纠正输入句子中的语法错误。"} \\&
 \texttt{"role": "user", "content": '请纠正下列向子中的语法错误，并确保纠正后向子的改动是最少的。请只输出修改后的语句，不要加入任何其他文字。开始:'}\\ 
\bottomrule
\end{tabular}}
\linespread{1}
\label{Tab:GEC_zero_shot_prompts}
\end{table}
\end{CJK*}

\subsection{Experimental Setup} 
To verify its effectiveness, we apply our CL method to three models: BART, mT5, and SynGEC. During the training phase, we adopt some hyper-parameters of SynGEC to guide our model's optimization process~\cite{zhang2022syngec}. We maintain the learning rate $5e^{-5}$ with a batch size of 8,192 tokens and fine-tune the parameters with Adam optimizer. Q, the number of Monte Carlo dropout samples, is set to 5.
In addition,  we carefully designed the prompts in Table~\ref{Tab:GEC_zero_shot_prompts} for evaluating the performance of LLM on existing exams. The temperature of LLM is set to 1.0. 

\begin{table}[ht]
\centering
\small
\caption{The performance of \methodName{} and all baselines on NLPCC-test dataset.}
\begin{tabular}{l|ccc|} 
\toprule
 Model& \textbf{Pre} & \textbf{Rec} & $\textbf{F}_{0.5}$ \\ 
\midrule
HRG~\cite{hinson-etal-2020-heterogeneous} & 36.79 & 27.82 & 34.56 \\
MaskGEC~\cite{zhao2020maskgec} & 44.36 & 22.18 & 36.97  \\
S2A~\cite{li2022sequence} & 42.34 & 27.11 & 38.06 \\ 
POL~\cite{wu-etal-2022-position} & 46.45 & 23.68 & 38.95 \\ 
SG-CGEC~\cite{wu-wu-2022-spelling}& 50.56 & 25.24 & 42.11 \\ 
TemplateGEC~\cite{li-etal-2023-templategec} & \textbf{54.50} & 27.40 & 45.50 \\
GPT-3.5-Turbo~\cite{ouyang2022training} &24.84 &{39.10} &26.80 \\
GPT-4~\cite{achiam2023gpt} &28.57 &\textbf{42.45} &30.57 \\
\midrule

mT5~\cite{xue-etal-2021-mt5}  & 47.16 & 31.27 & 42.81 \\
\methodName{} (mT5) & $\text{48.52}$ & 31.84 & $\text{43.92}$\\
\midrule
BART~\cite{shao2021cpt} & 49.07 & 32.80 & 44.64\\
\methodName{} (BART) & $\text{51.84}$ & 31.26 & $45.81$ \\ 
\midrule
SynGEC~\cite{zhang2022mucgec}  & 49.96 & 33.04 & 45.32 \\
\methodName{} (SynGEC) & $\text{52.05}$ & 31.52 & $\textbf{46.05}$\\
\bottomrule
\end{tabular}

\label{main_results_cgec_1}
\end{table}
\begin{table}[ht]
\centering
\small
\caption{The performance of \methodName{} and all baselines on MuCGEC-dev dataset. }
\begin{tabular}{l|ccc|} 
\toprule
 Model& \textbf{Pre} & \textbf{Rec} & $\textbf{F}_{0.5}$ \\ 
\midrule
mT5~\cite{xue-etal-2021-mt5}  & 41.57 & 25.38 & 36.87 \\
\methodName{} (mT5) & $\text{43.66}$ & 25.67 & $\text{37.42}$\\
\midrule
BART~\cite{shao2021cpt} & 42.02 & \textbf{26.45} & 37.62\\
\methodName{} (BART) & $\text{43.36}$ & 26.80 & $\text{38.59}$ \\ 
\midrule
SynGEC~\cite{zhang2022mucgec}  & 44.60 & 24.51 & 38.31 \\
\methodName{} (SynGEC) & $\textbf{45.94}$ & 24.66 & $\textbf{39.18}$\\
\bottomrule
\end{tabular}

\label{main_results_cgec_2}
\end{table}
\subsection{Main Results}
Table~\ref{main_results_cgec_1} and Table~\ref{main_results_cgec_2} illustrate the performance of our CL method when compared to the baselines.
From the experimental results, it is evident that our method achieves consistent improvements with a significant margin over all baselines by reordering the training data and focusing on hard instances. Our Curriculum Learning strategy forces the CGEC models to shift the emphasis toward more difficult samples gradually, which leads to an enhancement in their performance. The large improvements observed from three models (i.e., BART, mT5, and SynGEC) demonstrate the model-agnostic characteristic of our proposed method. 

\subsection{Ablation Study}
\label{sec:ablation}

To verify the effectiveness of each module in our proposed CL method, we conduct ablation studies with the following settings: 1) \textbf{Vanilla BART}, 2) \textbf{Only Batch-Level CL}, 3) \textbf{Only Instance-Level CL}. 
The model performance decreases when Batch-Level CL or Instance-Level CL is removed, which demonstrates the efficacy of each module in our proposed Loss-Aware Curriculum Learning framework.

From Table~\ref{ablation_results}, we can observe that each component of our approach is proven to produce a significant improvement to the model compared to Vanilla BART. Batch-Level CL forces the CGEC models to shift the emphasis toward the complex samples gradually, which leads to an enhancement in their performance. Instance-Level CL helps the CGEC model automatically optimize in the appropriate direction, which boosts its performance.

\begin{table}[t]
    \centering
    \small
        \caption{Results for ablation studies. "$\Delta$" indicates the absolute $F_{0.5}$ improvements on NLPCC-test dataset.}
    \begin{tabular}{l|c|c}
        \hline
          \textbf{Method} & $F_{0.5}$ & $\Delta$\\
         \hline
         BART & 44.64\% & -- \\
         +Only Batch-Level CL & 45.02\% & +0.38\% \\
         +Only Instance-Level CL & 45.28\% & +0.64\% \\
        \hline
         CL (BART) & \textbf{45.81\%} & \textbf{+1.17\%} \\
         \hline
    \end{tabular}
    \label{ablation_results}
    \vspace{-0.1cm}
\end{table}

\subsection{Parameter Study}
As mentioned in Sec~\ref{sec:batch_cl}, we assign k to scale the gap between the high-difficulty sentence and the low-difficulty sentence. In this section, to investigate the impact of different k, we perform extensive experiments by varying the valuation of k on the NLPCC-test dataset. Table~\ref{Ablation}{} shows that when the value of $k$ reaches a certain value, the performance of the model does not improve anymore. We believe that the overlarge k leads to overfit on hard instances and ignore easy instances. Consequently, it is essential to choose the optimal value of $k$, although there are consistent improvements when using BART as the PLM at all values of $k$ from 1 to 5.

\begin{table}[h]
\small
\centering
\caption{The $F_{0.5}$ scores on NLPCC-test and MuCGEC-dev, using different values of k in CL (BART).}
\begin{tabular}{lcrr}
\hline Model &k Value& NLPCC & MuCGEC \\
\hline BART &--& 44.64 & $37.62$ \\
CL (BART)&$k=1$ & 45.55 & $38.27$ \\
CL (BART)&$k=2$ & 45.81 & $38.59$ \\
CL (BART)&$k=3$ & 45.72 & $38.51$ \\
CL (BART)&$k=4$ & 45.65 & $38.44$ \\
CL (BART)&$k=5$ & 45.58 & $38.36$ \\
\hline
\end{tabular}

\label{Ablation}
\end{table}

\begin{figure}[]
\centering
\includegraphics[width=0.38\textwidth]{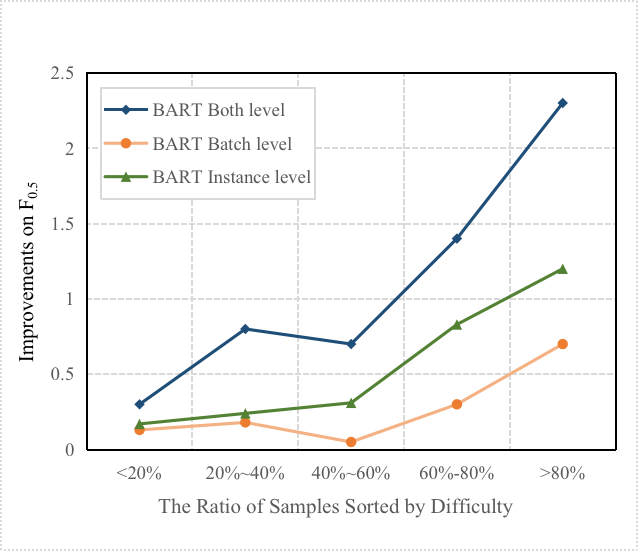}
\caption{Improvements of $F_{0.5}$ at different difficulty intervals on NLPCC-test.}
\label{correlation}
\end{figure}

\subsection{Correlation Between the Curriculum Learning and Improvements}
Despite the fact that our methods yield improvements across all the sentence pairs, the reasons behind the improvements remain unclear. Figure\ref{correlation} shows the $F_{0.5}$ improvements at different difficulty intervals on the NLPCC-test dataset.
The difficulty scores are calculated by the formula detailed in section ~\ref{sec:batch_cl}.
We observe that our approach consistently outperforms the BART baseline across various difficulty intervals. Both easy and difficult sentences benefit from the CL framework. The hardest sentences have notable improvements, which can be attributed to the emphasis on the hard parts during the training stage. 

\section{Conclusion}\label{formats}
In this paper, we propose a multi-granularity Loss-Aware Curriculum Learning method to enhance the CGEC models. Specifically, a novel loss-aware difficulty criteria is first proposed to assist the CGEC model in learning from simple samples to complex samples at Batch-Level. Then the difficulty scores of each instance at Sentence-Level and Token-Level are calculated to regulate the loss function, automatically facilitating the optimization of CGEC models in the appropriate direction. 
In conclusion, extensive experiments indicate that our CL method at different levels can effectively enhance the correction performance of CGEC models.

\bibliographystyle{IEEEtran}
\bibliography{reference} 


\end{document}